\pdfoutput=1

\documentclass[11pt]{article}

\usepackage[final]{acl}
\usepackage{footmisc}
\usepackage{times}
\usepackage{latexsym}
\usepackage{amssymb}
\usepackage {amsmath}
\usepackage{xcolor}
\usepackage{booktabs} 
\usepackage{array} 
\usepackage{graphicx}
\usepackage{booktabs}
\usepackage{multirow}
\usepackage[utf8]{inputenc}
\usepackage[most]{tcolorbox}
\usepackage{tikz}
\usepackage[T1]{fontenc}

\usepackage[utf8]{inputenc}

\usepackage{microtype}

\usepackage{inconsolata}

\usepackage{graphicx}
\usepackage{enumitem}

\setenumerate[1]{itemsep=0pt, partopsep=0pt, parsep=\parskip, topsep=5pt}
\setitemize[1]{itemsep=0pt, partopsep=0pt, parsep=\parskip, topsep=5pt}
\setdescription[1]{itemsep=0pt, partopsep=0pt, parsep=\parskip, topsep=5pt}

\usepackage[most]{tcolorbox}  
\definecolor{darkgrey}{rgb}{0.53,0.53,0.53}
\definecolor{mygrey}{rgb}{0.9,0.9,0.9}
\definecolor{purple}{RGB}{230, 227, 254}
\definecolor{lightgreen}{RGB}{238, 252, 241}
\definecolor{lightred}{RGB}{231, 187, 187}
\definecolor{darkred}{RGB}{198, 129, 129}

\definecolor{tabhighlight}{HTML}{e5e5e5}
\definecolor{someorange}{rgb}{0.773,0.353,0.067}
\definecolor{someblue}{rgb}{0.27, 0.35, 0.760}

\title{CAT: Causal Attention Tuning For Injecting Fine-grained Causal Knowledge into Large Language Models}

\author{
 \textbf{Kairong Han\textsuperscript{1}},
 \textbf{Wenshuo Zhao\textsuperscript{1}},
 \textbf{Ziyu Zhao\textsuperscript{1}},
 \textbf{Junjian Ye\textsuperscript{2}},
 \textbf{Lujia Pan\textsuperscript{2}},
 \textbf{Kun Kuang\textsuperscript{1\dag}}
\\
 \textsuperscript{1}College of Computer Science and Technology, Zhejiang University,\\
 \textsuperscript{2}Noah’s Ark Lab, Huawei Technologies,\\
 zju\_handso@163.com, \{zhao\_ws, kunkuang\}@zju.edu.cn,\\benzhao.styx@gmail.com, \{yejunjian, panlujia\}@huawei.com
}
\begin{document}
\maketitle
\begin{abstract}
Large Language Models (LLMs) have achieved remarkable success across various domains. However, a fundamental question remains: Can LLMs effectively utilize causal knowledge for prediction and generation? Through empirical studies, we find that LLMs trained directly on large-scale data often capture spurious correlations rather than true causal relationships, leading to suboptimal performance, especially in out-of-distribution (OOD) scenarios. 
To address this challenge, we propose Causal Attention Tuning (CAT), a novel approach that injects fine-grained causal knowledge into the attention mechanism.
We propose an automated pipeline that leverages human priors to automatically generate token-level causal signals and introduce the Re-Attention mechanism to guide training, helping the model focus on causal structures while mitigating noise and biases in attention scores. Experimental results on our proposed Spurious Token Game (STG) benchmark and multiple downstream tasks demonstrate that our approach effectively leverages causal knowledge for prediction and remains robust in OOD scenarios. The CAT achieves an average improvement of 5.76\% on the STG dataset and 1.56\% on downstream tasks. Notably, the OOD performance of the Llama-3.1-8B model on STG\_M increased from 64.5\% to 90.5\%, and Qwen’s OOD performance on the STG\_H dataset improved from 25.4\% to 55.9\%. Implementation details can be found ~\href{https://github.com/Kairong-Han/CAT}{here}.

\end{abstract}
\renewcommand{\thefootnote}{\fnsymbol{footnote}}
\footnotetext[0]{\dag\ Corresponding author.}

\section{Introduction}

\begin{figure}[ht]
    \centering
    \includegraphics[width=1\linewidth]{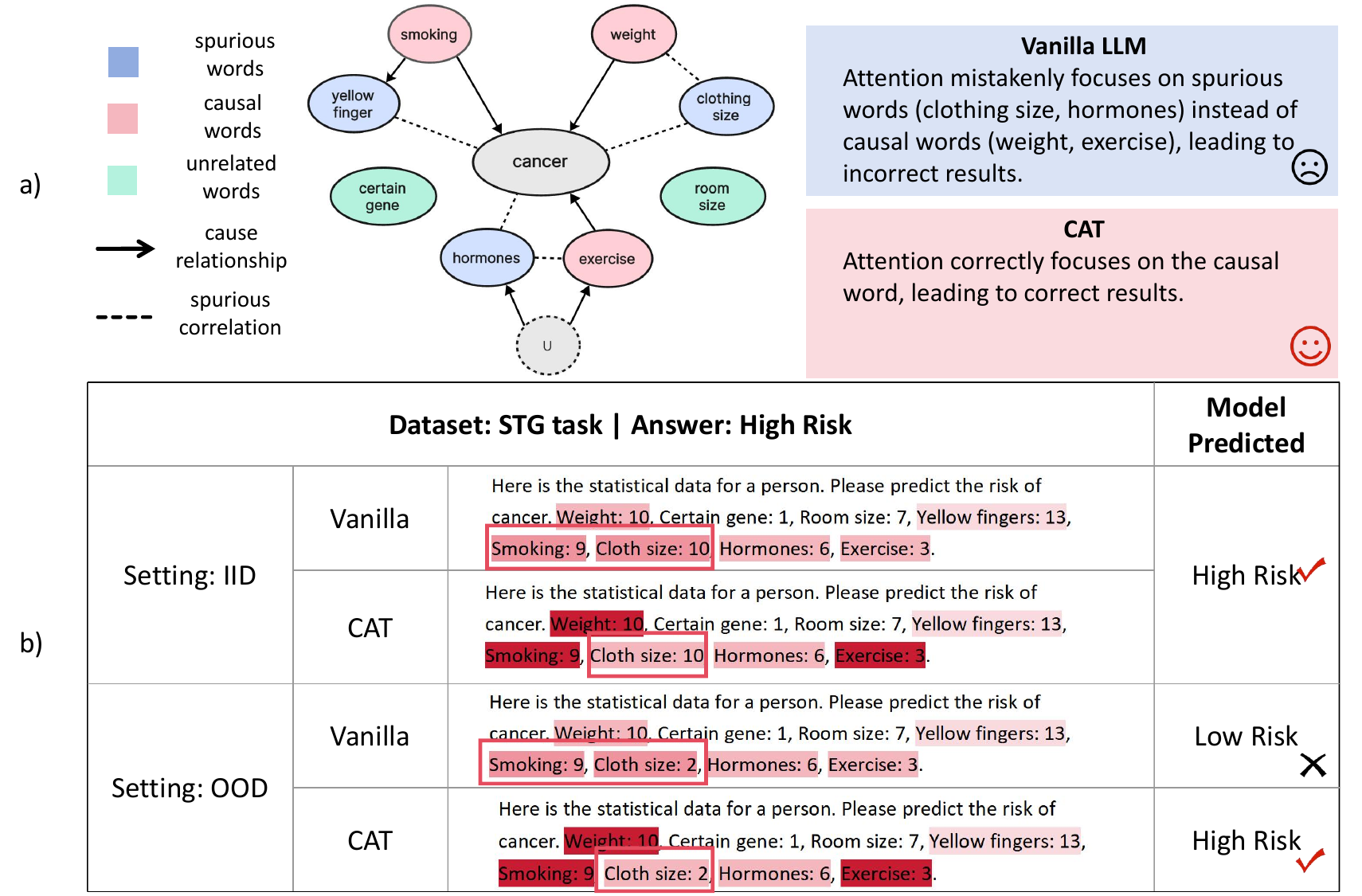}
    \caption{a) Training data is generated from this causal graph. LLM is influenced by spurious correlations and fails to learn causal relationships. b) The visualization of the attention distribution, where the deeper the red color, the higher the value. After training, the vanilla LLM incorrectly attends to spurious factors (e.g., clothing size), leading to failure in OOD scenarios. The CAT method, by injecting fine-grained causal knowledge, demonstrates stronger robustness in OOD scenarios.}\label{fig:intro3}
\end{figure}

Large Language Models (LLMs), trained in an autoregressive manner and guided by scaling laws, have achieved remarkable results across various domains~\cite{zhao2023survey,hadi2023survey,zhou2024comprehensive,huang2022towards,wang2024survey,Hu_2025_CVPR}. Their foundational architecture, the Transformer~\cite{vaswani2017attention}, leverages the attention mechanism to capture token-level correlations, which is central to their success. However, existing fine-tuning paradigms primarily focus on aligning LLMs with task-specific objectives. Relying solely on superficial correlations in the data can lead to spurious correlations, causing biases and negatively impacting its reasoning ability and generalization~\cite{wu2024causality,zhang2024causal}, which raises a critical question: \textit{Can LLMs truly learn and utilize causal relationships, rather than merely modeling surface-level correlations?}

To investigate this issue, we constructed a Spurious Token Game (STG) dataset, where the series attributes are classified into three types: causal factors, spurious factors\cite{ye2024spurious,dongerict}, and irrelevant factors. As illustrated in Figure~\ref{fig:intro3}, the numerical value of the spurious factor is proportional to the numerical value of the corresponding causal factor (i.e. cloth size has the same value as weight, 10 in the Figure).  However, in the OOD~\cite{liu2021towards,tong25latent,HTCL} test, this association pattern is removed to verify whether the model can use causal knowledge for prediction (i.e., a change in clothing size to 2 does not affect cancer risk), more details in Appendix \ref{DGP}. Our observations indicate that after direct fine-tuning, LLMs inherently allocate equal attention to both spurious and causal words, resulting in poor generalization in OOD scenarios~\cite{dai2024bias,gallegos2024bias}. This suggests that direct fine-tuning often leads models to prioritize spurious correlations over genuine causal relationships, ultimately impairing their generalization ability. Consequently, this raises concerns about their robustness, as LLMs driven by associative learning tend to internalize dataset biases, making them less reliable in handling diverse and unpredictable real-world scenarios.

To solve this problem, we propose the \underline{\textbf{C}}ausal \underline{\textbf{A}}ttention \underline{\textbf{T}}uning (CAT), a novel approach that integrates fine-grained causal knowledge into the attention mechanism. Specifically, the method consists of two steps. First, to automate the generation of causal supervision signals, human experts manually write a few examples and leverage an assistant LLM to generate causal supervision signals for a large-scale dataset. Second, to embed causal knowledge into attention, we convert word-level supervision signals into an adjacency matrix, aligning it with the attention training objective. Then, we introduce the Re-Attention mechanism, which guides model training by constraining the average attention map based on causal prior knowledge. Through the above approach, we align the decision-making process of LLMs with human causal knowledge at the attention level, effectively intervening in the model’s decision dependencies. Finally, we develop a new benchmark STG to systematically evaluate whether LLMs can capture causal knowledge.

Experiments on STG demonstrate that CAT effectively directs the model's attention toward causal features, leading to consistent improvements in both independent and identically distributed (IID) and OOD settings. For example, the OOD performance of the Llama-3.1-8B model on STG\_M increased from 64.5\% to 90.5\%, and Qwen’s OOD performance on the STG\_H dataset improved from 25.4\% to 55.9\%. Furthermore, evaluations on five widely used mathematical and reasoning tasks show that incorporating causal knowledge through CAT enhances the performance on downstream tasks. The CAT can be seamlessly integrated with mainstream training methods such as LoRA\cite{hu2021lora}, demonstrating its general applicability. Our contributions are summarized as follows:

\vspace{-1.5mm}
\begin{itemize}[leftmargin=*]
\setlength{\parskip}{3pt}
    \item  We constructed a new benchmark called STG to evaluate whether LLMs can capture and use causal knowledge.
    
    \item We propose the CAT, a novel approach for integrating causal knowledge into LLMs. Through the Re-Attention mechanism, we mitigate noise and bias in the attention mechanism, resulting in performance improvements in both IID and OOD scenarios.

    \item Experiments on STG and mathematical and reasoning downstream tasks demonstrate the strong generalization ability of the CAT. The CAT achieves an average improvement of 5.76\% on the STG dataset and 1.56\% on downstream tasks. Notably, the OOD performance of the Llama-3.1-8B model on STG\_M increased from 64.5\% to 90.5\%, and Qwen’s OOD performance on the STG\_H dataset improved from 25.4\% to 55.9\%, demonstrating the potential of the Re-Attention mechanism
\end{itemize}

\section{Related Work}

\subsection{Combine Causality and LLMs}
Since the advent of LLMs, researchers have explored ways to enhance their capabilities by integrating causal theory with LLMs~\cite{wu2024causality,han2024causal}.

In the areas of debiasing and fairness \cite{meade2021empirical,wang2024mementos}, Counterfactual Data Augmentation (CDA) \cite{webster2020measuring} is proposed to solve gender bias, which generates counterfactual samples by flipping gender-related keywords. Building on counterfactual generation, invariant loss \cite{zhou2023causal} is introduced to further mitigate biases related to gender and other stereotypes. Entity bias \cite{longpre2021entity} is another kind of bias, and "do" operations \cite{pearl2010causal} on intermediate variables of both white-box and black-box large language models \cite{wang-etal-2023-causal} is proposed to eliminate it. Jenny et al. employed activity dependency networks to better explain bias perspectives that were previously simplified only through correlations \cite{jenny2024exploring}. Zhou et al focused on conceptual bias and used counterfactual data generated by ChatGPT to balance label distributions and mitigate spurious correlations \cite{zhou2023explore}. Wu et al. propose the De-biased Attention Supervision (DAS) \cite{Wu_Liu_Zhao_Lu_Zhang_Sun_Wu_Kuang_2024} method, using the backdoor adjustment to mitigate bias caused by the label distribution of the dataset.  However, the aforementioned debiasing works are limited to specific scenarios and have limited practical applicability. In terms of reasoning capabilities, causal prompt \cite{zhang2024causal} is proposed, leveraging Chain-of-Thought (COT) for front-door adjustment to enhance reasoning performance, but the reasoning overhead is significant. Bao et al. and Li et al. addressed the issue of unfaithful COT by modeling and mitigating it from a causal explanation perspective \cite{bao2024llms,li2024towards}. Jin et al. introduced CausalCOT \cite{jin2023cladder}  to enhance causal reasoning abilities. Feng et al focused on learning a robust classifier across multiple domains \cite{feng-etal-2024-imo}. 

Different from the aforementioned works, the CAT injects causal prior knowledge into the attention training process, offering a simple and efficient method with strong generalization ability for prediction and generation.

\subsection{Research on Attention Mechanism}
LLMs have sparked new explorations into the Attention mechanism. A series of studies focused on attention mechanisms \cite{niu2021review,guo2022attention,hu2025dfusiondirectpreferenceoptimization}, aiming to explain and intervene in attention score distribution. Research has shown that the attention mechanism can extract reasonable word alignments, with attention scores and their norms collectively determining the output \cite{kobayashi-etal-2020-attention}. However, attention often allocates a significant portion of focus to tokens with no semantic value, a phenomenon termed "attention sinks" 
 \cite{sun2024massive,gu2024attention}, which has been utilized to enhance long-context outputs 
 \cite{xiao2023efficient}. However, maintaining these attention sinks is not always beneficial, and researchers have observed consistent performance improvements across various models by redistributing excess attention scores to other tokens \cite{yuunveiling}. To optimize the attention distribution, the differential transformer \cite{ye2024differential}, inspired by signal denoising systems, adopts a sparse attention pattern, leading to performance improvements in LLMs.  Nevertheless, the aforementioned methods overlook the inherent biases and spurious correlations in the data.

The CAT introduces causal prior supervision signals into attention training. By injecting fine-grained causal knowledge into the attention mechanism, we aim to accomplish debiasing and denoising at the architectural level, rather than merely fitting the data distribution of downstream tasks.

\section{Preliminary}\label{Preliminary Concepts}
\subsection{Attention Mechanism}

We start with the vanilla Transformer, where an input sequence is mapped to a feature matrix $\mathbf{S} = [\mathbf{s}_1, \dots, \mathbf{s}_n]^\top \in \mathbb{R}^{n \times d_{\text{model}}}$ through vocabulary embedding and positional encoding. Each of the $n$ tokens is represented by a $d_{\text{model}}$-dimensional vector. We focus on the attention mechanism, which models dependencies between tokens:
$$ \mathbf{Q}_i,\mathbf{K}_i,\mathbf{V}_i=\mathbf{S} \cdot 
 \mathbf{W}_i^Q,\mathbf{S} \cdot \mathbf{W}_i^K,\mathbf{S} \cdot \mathbf{W}_i^V,$$
$$\textbf{Z}_i^{attn} =  \underset{\text{attention map}}{\underbrace{\text{softmax}\left(\frac{\mathbf{Q}_i \cdot \mathbf{K}_i^\top}{\sqrt{ d_k}}\right)} \cdot \mathbf{V}_i},
$$
$$ \textbf{Z}_{mult}=\text{Concat}(\textbf{Z}^{attn}_1,\dots,\textbf{Z}^{attn}_h)\cdot \mathbf{W}_O,$$
where for each head $i$ in multi-head attention, $W^Q_i \in \mathbb{R}^{d_{\text{model}} \times d_k}$, $W^K_i \in \mathbb{R}^{d_{\text{model}} \times d_k}$, $ W^V_i \in \mathbb{R}^{d_{\text{model}} \times d_v}$, $W_O \in \mathbb{R}^{hd_v \times d_{\text{model}}}$.

\subsection{Causal Graph}
Causal graph is the structured representation of causal knowledge \cite{lipsky2022causal,thulasiraman2011graphs}, denoted as a directed acyclic graph (DAG) \(G = \{<V^G, E^G>\}\), where a directed edge \(v_i \rightarrow v_j \in E^G\) indicates that element \(v_i \in V^G\) is the direct cause of element \(v_j \in V^G\), i.e., \(v_i\) causes \(v_j\). We use the adjacency matrix corresponding to the causal graph DAG to align the training objective of attention.

\begin{figure*}[ht]
    \centering
    \includegraphics[width=1\linewidth]{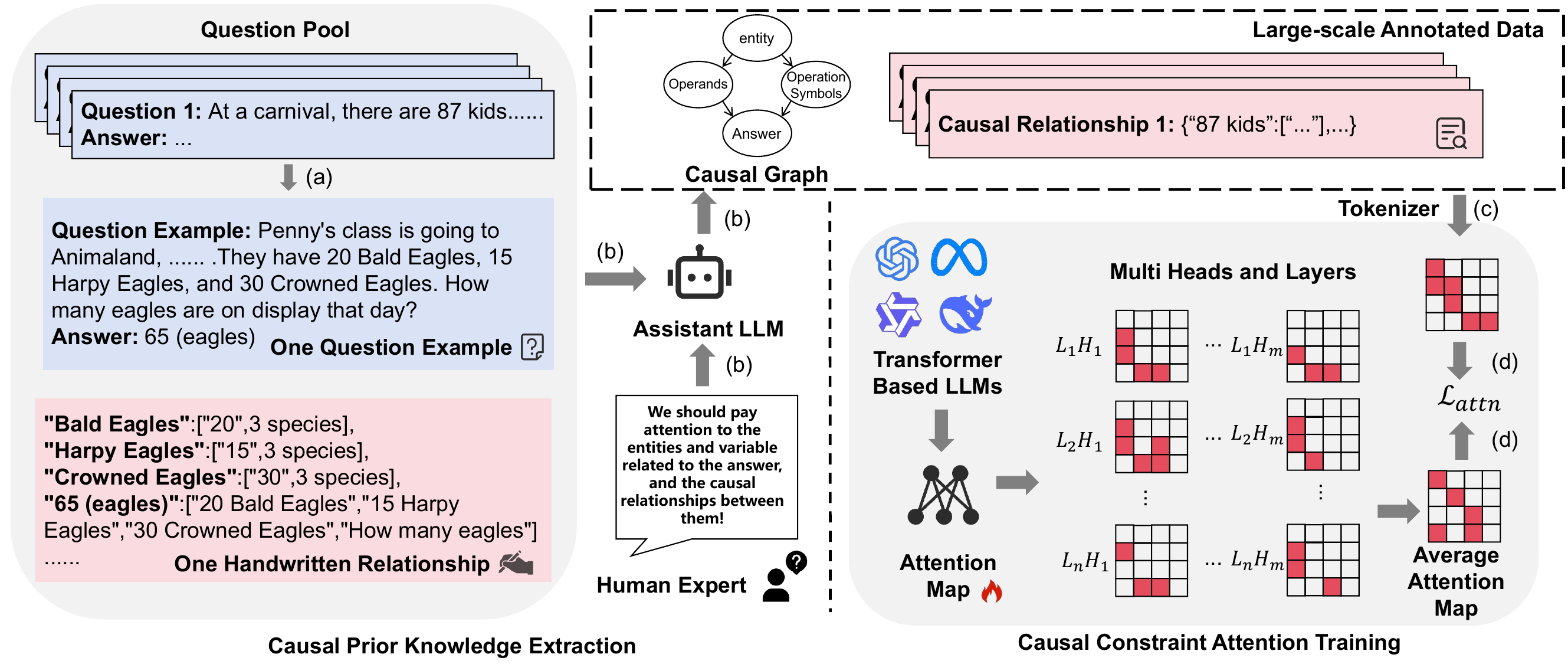}
    \caption{(a) Human experts construct handwritten causal relationships at the word level for downstream tasks. (b) The assistant LLM automates the annotation of downstream tasks based on handwritten examples. (c) Token-level causal associations are obtained using the tokenizer and transformed into an adjacency matrix. (d) The Re-Attention mechanism is employed to train LLMs by introducing $\mathcal{L}_{attn}$, which injects fine-grained parameterized causal knowledge to intervene the model’s decision dependencies.}
    \label{fig:method}
\end{figure*}

\section{Methodologies}
In this section, we introduce the CAT framework, as shown in Figure~\ref{fig:method}. The CAT comprises two key steps: (1) causal prior knowledge extraction and (2) causal constraint attention training.

\subsection{Causal Prior Knowledge Extraction} \label{Causal Prior Knowledge Extraction}

Due to the inherent complexity of natural language, aligning token-level causal relationships with existing causal prior knowledge presents three challenges:
\begin{itemize}
    \item  Causal relationships in natural language text are difficult to simply identify using rule-based matching.
    \item The specific design of tokenizers can lead to the fragmentation of a single word into multiple tokens, adding complexity for LLMs in effectively incorporating causal knowledge. 
    \item  The high cost of large-scale annotation by human experts hinders scalability.
\end{itemize}
  To address these challenges, we propose an automated pipeline for the generation of causal supervision signals.

\textbf{Step 1: Prompt generation.} Even though downstream tasks may involve complex and diverse expressions, the causal words that ultimately lead to the answer are considered to carry rich semantic information. In mathematical reasoning, we focus on numerical values, entities, numerical operation symbols, and the causal relationships between these words. Following this heuristic, we construct a prompt to guide the assistant LLM in uncovering causal relationships for the downstream task. The prompt consists of a task description \(\mathcal{P}_t\) and handwritten specific examples \(\mathcal{P}_d\), i.e., 65 eagles are calculated by 20 Bald, 15 Harpy, and 30 Crowned in Figure \ref{fig:method}\textbf{(left)}. Although the causal graph is not explicitly provided, the handwritten examples follow the causal logic used by humans to solve the problem. The detailed prompt templates can be found in Appendix \ref{Prompt Template}.

\textbf{Step 2: Token-Level causal knowledge extraction.} Using large-scale human expert annotations is cost-intensive. so we annotate the training data by inputting \(\mathcal{P}_t\) and \(\mathcal{P}_d\) from Step 1, along with the downstream task question description \(\mathcal{Q}\) and answer \(\mathcal{A}\), into assistant LLM to obtain textual supervision signals \(\mathcal{M}\):  
\[  
\mathcal{M} = \text{LLM}([\mathcal{P}_t; \mathcal{P}_d; \mathcal{Q}; \mathcal{A}])  .
\]  
To obtain a structured representation of causal relationships, we constrain the textual supervision signals to be in JSON format as a dictionary \(\mathcal{M} = \{(k^M, v^M)| k^M \in \mathcal{Q}\bigcup \mathcal{A},v^M \in \mathcal{Q}\bigcup \mathcal{A}\}\). This implies the generation of the \(k^M\) is primarily influenced by the \(v^M\). Based on the specific tokenizer implementation, we convert the textual supervision signals into an adjacency matrix \(\mathbf{A}^{adj} \in \{0, 1\}^{n \times n} \):  
\[
\mathbf{A}^{adj}_{i,j} = 
\begin{cases} 
1, & (token(i), token(j)) \in \mathcal{M} \\
0, & \text{else}
\end{cases}
\]
where \(\mathbf{A}^{adj}_{i, j} = 1\) indicates that the \(i\)-th token causes the \(j\)-th token, while \(\mathbf{A}^{adj}_{i, j} = 0\) indicates no causal relationship. $(token(i),token(j))$ means the tuple formed by the words corresponding to the i-th and j-th tokens. Tokens identified as having causal relationships require additional attention during training.

\subsection{Causal Constraint Attention Training}\label{Causal Constraint Attention Training}

The attention map can be re-written in the following matrix form:

\begin{center}
\begin{equation}
\label{eq:Transformer}
\begin{split}
 \textbf{Z}_i^{attn} &=\frac{1}{\sqrt{ d_k}}\begin{bmatrix}
a_{1,1} & 0 & \cdots & 0 \\
a_{2,1} & a_{2,2} & \cdots & 0 \\
\vdots & \vdots & \ddots & \vdots \\
a_{n,1} & a_{n,2} & \cdots & a_{n,n}
\end{bmatrix}
\begin{bmatrix}
v_1\\v_2\\\vdots \\ v_n
\end{bmatrix}, \nonumber
\end{split} 
\end{equation}
\end{center}
where $a_{i,j}$ denote the dot product $q_i \cdot k_j^\top$, and $q_i$ is the query from the $i$-th token and $k_j$ is the key from the $j$-th token. In next-token prediction, the $i$-th token serves as input for generating the $(i+1)$-th token. The model assigns attention weights to the first $i$ tokens based on the $i$-th row of the attention map, where each $a_{i,j}$ reflects the importance of token $j$ in predicting token $(i+1)$. To inject causal prior knowledge, we encourage the model to focus more on tokens that are causally related to the one being generated. This is achieved by shifting the token-level causal adjacency matrix $\mathbf{A}^{\text{adj}}$ upward by one position, aligning it with the next-token generation process:
$$\mathbf{A}^{adj}_{i, j} = \mathbf{A}^{adj}_{i+1,j}.$$
In multi-head attention, due to the difficulty in precisely quantifying the importance of different layers and different heads within each layer for downstream tasks, we consider the average attention map $\overline{\mathbf{A}^M}$ across all layers $L$ and all heads $H$:
$$\overline{\mathbf{A}^M}=\frac{1}{L*H}\sum_{l=1}^{L}\sum_{h=1}^{H} \text{softmax}\left(\frac{\mathbf{Q}_{l,h} \cdot \mathbf{K}_{l,h}^\top}{\sqrt{ d_k}}\right).$$
To enforce the attention mechanism to focus more on token-level causal relationships, we utilize the adjacency matrix of causal words from the previous section as a supervision signal. Specifically, in the average attention map \(\overline{\mathbf{A}^M}\), for rows \(i\) where causal words appear, we calculate the average attention score \(\mathcal{C}_i\) of the tokens corresponding to the causal words in that row. For the remaining tokens in the row, we compute the average attention score \(\mathcal{N}_i\):
$$\mathcal{C}_i =\frac{1}{\sum_{j=1}^i \mathbf{A}^{adj}_{i,j}} \sum_{j=1}^{i}\overline{\mathbf{A}^M_{i,j}}\cdot \mathbf{A}^{adj}_{i,j},$$
$$\mathcal{N}_i = \frac{1}{\sum_{j=1}^i(1-\mathbf{A}^{adj}_{i,j})}\sum_{j=1}^{i}\overline{\mathbf{A}^M}_{i,j}\cdot (1-\mathbf{A}^{adj}_{i,j}).$$
We aim to ensure that the attention score of causal tokens in each row is no less than \(\alpha\) times the average attention score of the remaining tokens. Therefore, we introduce the following loss:  
\[
\mathcal{L}_{attn} = \sum_{i=0}^{n}\max(0, \alpha - \frac{\mathcal{C}_i}{\mathcal{N}_i}).
\]
This process allows attention to refocus on the causal relationship between tokens, the so-called \textbf{\textit{Re-Attention}} mechanism, as shown in Figure \ref{fig:method}\textbf{(right)}. For both pre-training and supervised fine-tuning (SFT) of LLMs, researchers employ the next token prediction loss. Specifically, given a sequence of tokens \(x_1, x_2, \ldots, x_T\) \cite{achiam2023gpt,devlin2018bert}:
\[
\mathcal{L}_{next} = -\sum_{t=1}^{T-1} \log P(x_{t+1} \mid x_{\leq t}).
\]
Therefore, the total loss during training is:
$$\mathcal{L}_{total} =\mathcal{L}_{next} +\gamma\mathcal{L}_{attn},$$
where \(\gamma\) is used to modulate the gradient when applying constraints to the attention mechanism. 

\section{Experiments}
\subsection{Experimental Setup}

\textbf{Baseline.} We conduct experiments on TinyLlama-1.1B\footnote{https://huggingface.co/TinyLlama/TinyLlama\_v1.1} \cite{zhang2024tinyllama}, Qwen2.5-1.5B\footnote{https://huggingface.co/Qwen/Qwen2.5-1.5B} \cite{qwen2.5,qwen2} and Llama-3.1-8B-Instruct\footnote{https://huggingface.co/meta-llama/Llama-3.1-8B-Instruct} \cite{touvron2023llama}, and conduct experiments under both full-parameter fine-tuning and parameter-efficient fine-tuning using LoRA\cite{hu2021lora}. 

\textbf{Dataset.} First, we evaluate CAT on two subsets of the STG dataset: STG\_Easy (STG\_E) and STG\_Hard (STG\_H). Furthermore, we used the following commonly used datasets related to mathematical reasoning, choice questions, and logical reasoning: MAWPS \cite{koncel-kedziorski-etal-2016-mawps}, ASDiv \cite{miao-etal-2020-diverse}, GSM8K \cite{cobbe2021gsm8k}, ARC\_E \cite{clark2018think}, and SVAMP \cite{patel-etal-2021-nlp}. For the aforementioned data without a partitioned test set, we randomly split it into training, validation, and test sets with a ratio of 6:2:2.

\textbf{Implementation Details.} All experiments were conducted on the NVIDIA A100 40GB GPU. 
 To guarantee the fairness of the comparison, we ensured that all hyperparameters were consistent between the baseline and the CAT. A warm-up coefficient of 0.1 was coupled with a cosine learning rate schedule and the AdamW optimizer. \(\gamma = e^{-i}\) where $i$ denotes the current epoch number. Unless otherwise specified, we use a default learning rate of $5\times e^{-5}$ for full fine-tuning, and $1\times e^{-4}$ for LoRA fine-tuning. The default number of training epochs is 4 for downstream tasks and 6 for the STG dataset. We use ChatGLM-4-air \cite{glm2024chatglm} as the assistant LLM. Other hyperparameter details and experiment details are shown in Appendix \ref{APP:hyper}.

\subsection{Results}
\subsubsection{Spurious Tokens Game}\label{sec:stg}

The STG benchmark consists of two subsets: STG\_E and STG\_H. To investigate the impact of data volume on spurious correlations, STG\_E is further divided into three scales of training datasets: large (STG\_L), medium (STG\_M), and small (STG\_S), along with an IID testing set and an OOD testing set.
\begin{figure}[ht]
    \centering
    \includegraphics[width=1\linewidth]{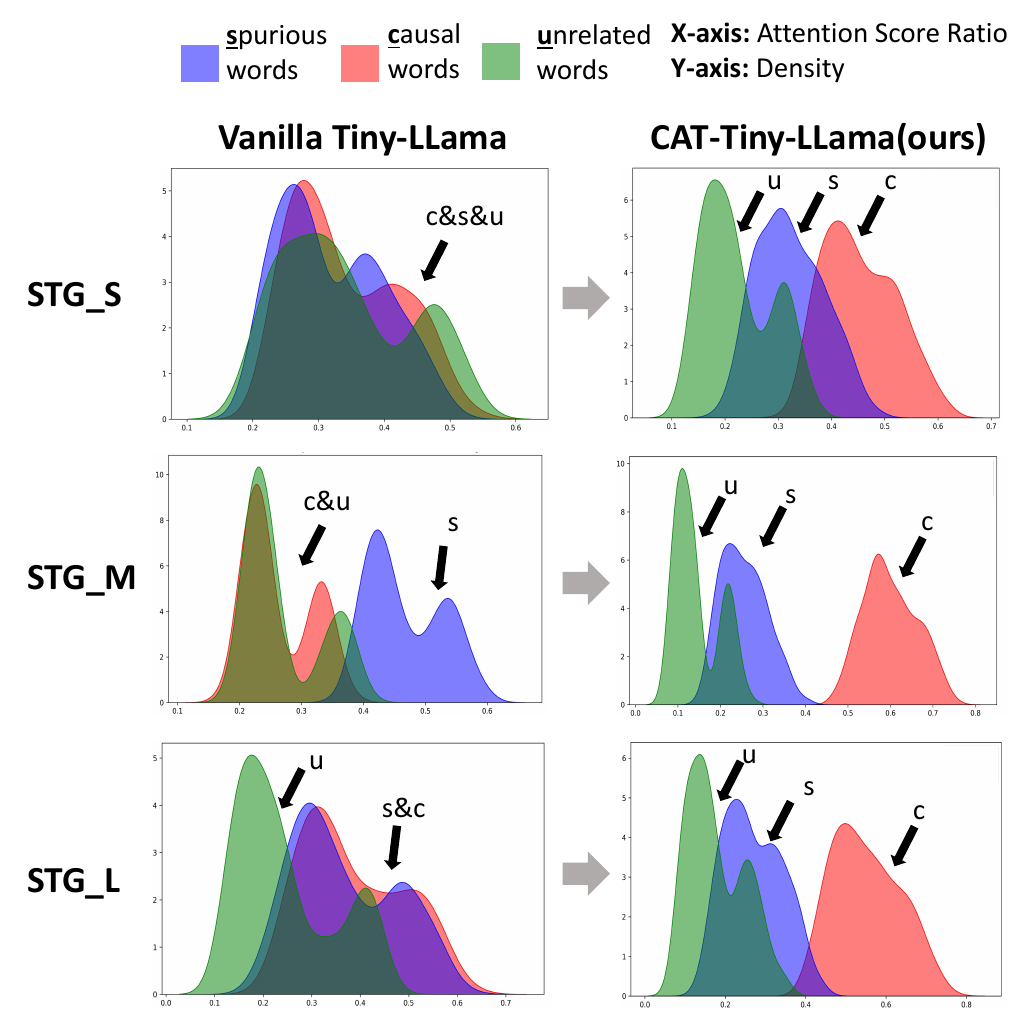}
    \caption{The density distribution of attention scores for three types of words in the average attention map under full parameters training when using TinyLlama-1.1B.}
    \label{fig:Density}
\end{figure}

In STG\_E, given a set of attributes and their corresponding values, the model learns to predict the risk of lung cancer. A specific example is illustrated in Figure \ref{fig:intro3}. Cancer risks are only caused by causal factors $\mathcal{C}^s$. Spurious factors $\mathcal{S}^s$ have a proportional relationship with causal factors. Irrelevant factors $\mathcal{I}^s$ are sampled independently from the previous two. In OOD tasks, we break the proportional relationship between spurious association factors and causal factors. In STG\_H, similar to STG\_E, the model is required to make predictions based on input variables. However, the key difference is that STG\_H includes a larger number of variables, and its answer is a continuous value ranging from 0 to 100, detailed in Appendix \ref{DGP}.  The experimental results are presented in Table \ref{stg}.
\begin{figure}[h]
    \centering
    \includegraphics[width=1\linewidth]{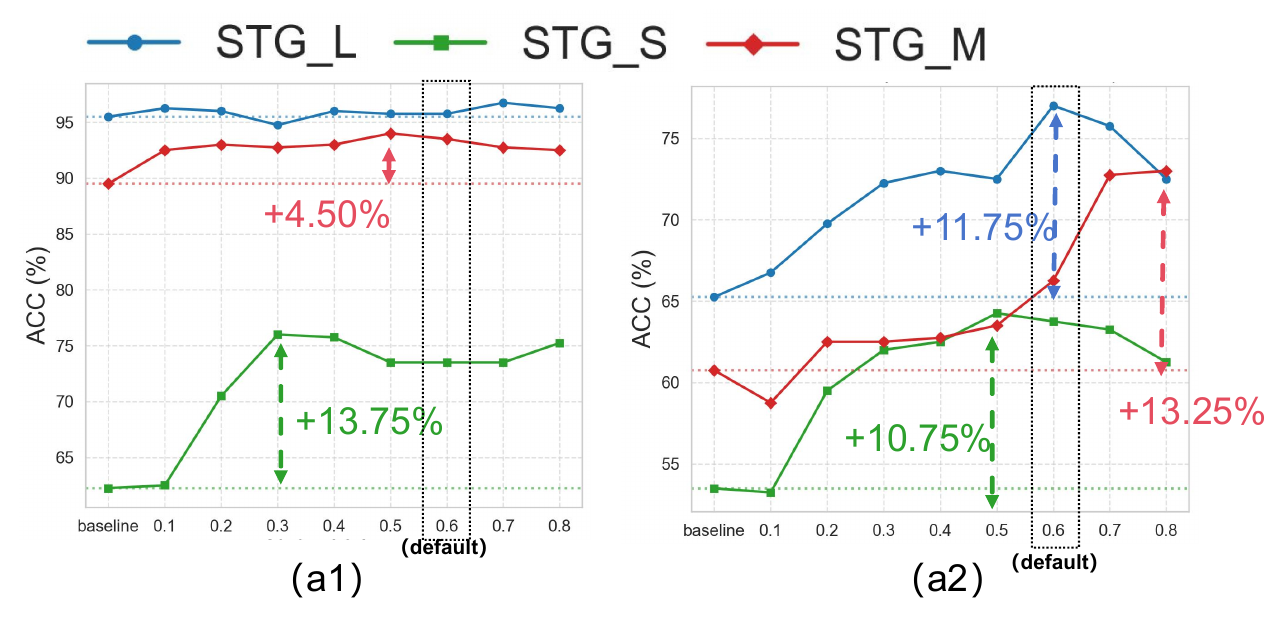}
    \caption{Results under full parameters training when using TinyLlama: a1) IID performance under different $\alpha$, a2) OOD performance under different $\alpha$.}
    \label{fig:iid}
\end{figure}

\begin{table*}[ht]
  \centering
  \resizebox{1\textwidth}{!}{ 
\begin{tabular}{cccc|cccc|c}
  \hline
  \multirow{2}{*}{\textbf{Model}} & 
  \multirow{2}{*}{\textbf{Setting}} & 
  \multirow{2}{*}{\textbf{Task}} & 
  \multirow{2}{*}{\textbf{Method}} &\multicolumn{3}{c}{\textbf{STG\_E}} & \multirow{2}{*}{\textbf{STG\_H}} & \multirow{2}{*}{\textbf{Average}} \\
  \cline{5-7}
   &  &  &  & \textbf{STG\_S} & \textbf{STG\_M} & \textbf{STG\_L} &  \\
  \hline
    \multirow{8}{*}{TinyLlama-1.1B}
    &\multirow{4}{*}{Full}
    &\multirow{2}{*}{IID}
    & \text{Vanilla} &62.25\%& 89.50\%&95.50\%&32.20\%&69.86\%\\ 
    &&& \text{CAT} &\textbf{73.50\%}& \textbf{93.50\%}& \textbf{95.75\%}&\textbf{37.10\%}&\textbf{74.96\%}\\ \cmidrule{3-9}
    &&\multirow{2}{*}{OOD}
    &\text{Vanilla} &53.50\%& 60.75\%& 65.25\%&4.10\%&45.90\%\\
    &&& \text{CAT} &\textbf{63.75\%}& \textbf{66.25\%}&\textbf{77.00\%}&\textbf{6.10\%}&\textbf{53.27\%}\\\cmidrule{2-9}

        &\multirow{4}{*}{LoRA}&\multirow{2}{*}{IID}
    &\text{Vanilla} &62.75\%& 83.50\%& \textbf{96.00\%}&31.90\%&68.54\%\\ 
    &&& \text{CAT} &\textbf{81.50\%}& \textbf{86.75\%}&\textbf{96.00\%}&\textbf{35.70\%}&\textbf{74.99\%}\\\cmidrule{3-9}
        &&\multirow{2}{*}{OOD}
    &\text{Vanilla} &59.25\%& 56.75\%& 61.50\%&5.90\%&45.85\%
\\
    &&& \text{CAT} &\textbf{65.50\%}& \textbf{63.50\%}&\textbf{69.50\%}&\textbf{9.30\%}&\textbf{51.95\%}\\
    \hline

     \multirow{8}{*}{Qwen2.5-1.5B}
    &\multirow{4}{*}{Full}
    &\multirow{2}{*}{IID}
    & \text{Vanilla} &55.00\%& \textbf{94.50\%}&95.50\%&56.60\%&75.40\%
\\ 
    &&& \text{CAT} &\textbf{74.00\%}& \textbf{94.50\%}& \textbf{95.75\%}&\textbf{62.50\%}&\textbf{81.69\%}\\ \cmidrule{3-9}
    &&\multirow{2}{*}{OOD}
    &\text{Vanilla} &53.50\%& \textbf{79.00\%}& 79.75\%&25.40\%&59.41\%\\
&&& \text{CAT} &\textbf{62.50\%}& \textbf{79.00\%}&\textbf{83.25\%}&\textbf{55.90\%}&\textbf{70.16\%}\\\cmidrule{2-9}

        &\multirow{4}{*}{LoRA}&\multirow{2}{*}{IID}
    &\text{Vanilla} &81.50\%& \textbf{93.25\%}& \textbf{95.75\%}&51.50\%&\textbf{80.50\%}
\\ 
    &&& \text{CAT} &\textbf{82.00\%}& 90.50\%&\textbf{95.75\%}&\textbf{53.40\%}&80.41\%\\\cmidrule{3-9}
        &&\multirow{2}{*}{OOD}
    &\text{Vanilla} &\textbf{78.50\%}& 82.00\%& 82.00\%&38.70\%&70.30\%
\\
    &&& \text{CAT} &\textbf{78.50\%}& \textbf{88.00\%}&\textbf{84.25\%}&\textbf{46.10\%}&\textbf{74.21\%}\\
    \hline

    \multirow{4}{*}{Llama-3.1-8B}
    &\multirow{4}{*}{LoRA}
    &\multirow{2}{*}{IID}
    & \text{Vanilla} &90.50\%& 93.25\%&96.00\%&57.80\%&84.39\%\\ 
    &&& \text{CAT} &\textbf{94.00\%}& \textbf{93.50\%}& \textbf{96.75\%}&\textbf{61.40\%}&\textbf{86.41\%}\\ \cmidrule{3-9}
    &&\multirow{2}{*}{OOD}
    &\text{Vanilla} &86.25\%& 64.50\%& 88.25\%&49.60\%&72.15\%\\
&&& \text{CAT} &\textbf{89.00\%}& \textbf{90.50\%}&\textbf{89.25\%}&\textbf{58.50\%}&\textbf{\textbf{81.81\%}}\\\hline
    
  \end{tabular}
  }
  \caption{Comparison of IID and OOD experimental results between Vanilla and CAT on the STG task under different settings. Full represents full parameters training, LoRA represents LoRA training.}\label{stg}
\end{table*}    

The CAT achieves significant improvements over the baseline in both IID and OOD scenarios. We take TinyLlama-1.1B as an example to explore the impact details of CAT on the attention mechanism. We visualize the distribution function of the average attention scores for three types of factors, as shown in Figure \ref{fig:Density}. Additionally, we also analyze the impact of different $\alpha$ values on IID and OOD generalization, as shown in Figure \ref{fig:iid}.

\textbf{Conclusion 1:} The distribution of attention scores for different tokens, obtained spontaneously and unsupervised, is difficult to predict and unstable. This means unclear decision dependencies.

As shown in Figure \ref{fig:Density}, with small data scale, the baseline exhibits similar attention distributions across all factor types, resulting in near-random (50\%) performance in both IID and OOD scenarios. With medium data, the baseline achieves 90\% accuracy in IID but relies on spurious correlations, leading to poor OOD performance. With large data, the attention score is similarly distributed between causal and spurious correlation factors. Additionally, spurious correlations have introduced significant instability. With the Llama-3.1-8B model, when the data size is doubled (from \_S to \_M), although IID performance continues to improve, the OOD performance actually drops from 86.25\% to 64.50\%. Moreover, in STG\_H, model performance consistently degrades greatly under OOD settings. In most settings, OOD performance is less than half of the IID performance. Simply scaling up the model size does not solve the problem. 

As a comparison, with the CAT method, the improvement is significant. On average, CAT improves the IID performance by 3.95\% and OOD performance by 7.56\%. Notably, the OOD performance of the Llama-3.1-8B model on STG\_M increased from 64.5\% to 90.5\%, and Qwen’s OOD performance on the STG\_H dataset improved from 25.40\% to 55.9\%.

\textbf{Conclusion 2:} As shown in Figure \ref{fig:iid}, within a certain range, the larger $\alpha$, the better the performance.

As $\alpha$ increases, the model’s attention to causal words grows. Across all dataset sizes, performance in both IID and OOD tasks initially improves. This suggests that greater attention to causal words enhances generalization. However, too much focus can disrupt the original attention distribution, causing a conflict with pre-trained parameters and resulting in performance degradation. 
More details in Appendix \ref{APP:STG}.

\textbf{Conclusion 3:} The CAT enhances the model's performance by mitigating noise and bias.

\begin{figure*}[h]
    \centering
    \includegraphics[width=1\linewidth]{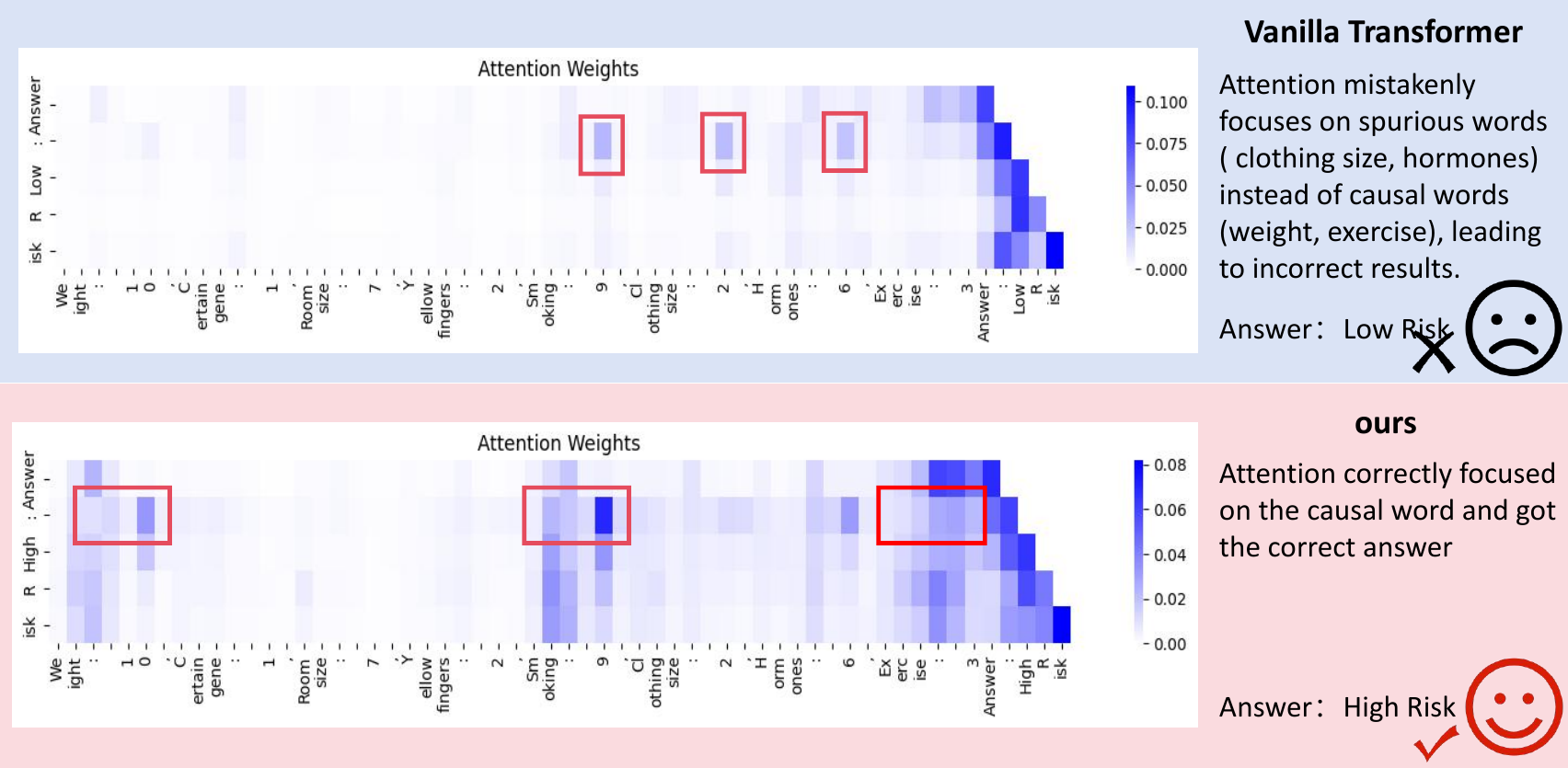}
    \caption{Visualization of the attention map in the STG task when using TinyLlama. }
    \label{fig:keshihua}
\end{figure*}

As shown in Figure \ref{fig:keshihua}, without causal priors, the model fails in OOD scenarios, focusing on spurious correlation factors instead of causal factors. In contrast, CAT enables the model to focus on causal factors. The CAT changes the model's decision dependency mechanism, making its performance more robust.
\begin{table*}[h!]
  \centering
  \resizebox{1\textwidth}{!}{ 
  \begin{tabular}{ccccccc@{\hskip 4pt}ccccc}
    \hline
    
    \multirow{2}{*}{\textbf{Model}} & 
  \multirow{2}{*}{\textbf{Setting}} & 
  \multirow{2}{*}{\textbf{Method}} & 
  \multicolumn{4}{c}{\textbf{In-Domain}} & \multicolumn{3}{c}{\textbf{Out-of-Domain}} & \multirow{2}{*}{\textbf{Average}} \\
  \cmidrule(r){4-7} \cmidrule(lr){8-10}
   &  &  & \textbf{MAWPS} & \textbf{SVAMP} & \textbf{ARC-E} & \textbf{GSM8K}   &\textbf{ASDiv}$^{\dag}$&\textbf{MAWPS}$^{\dag}$& \textbf{SVAMP}$^{\dag}$  \\
  \hline
    
    \multirow{6}{*}{TinyLlama-1.1B} & \multirow{3}{*}{Full} & Vanilla & 38.98\% & 17.50\% & 25.38\% & 13.04\% & 13.14\% & 20.68\% & 11.00\% & 19.96\% \\
    & & CAT & \textbf{41.16\%} & \textbf{20.00\%} & \textbf{26.01\%} & \textbf{14.18\%} & \textbf{14.04\%} & \textbf{23.15\%} & \textbf{11.40\%} & \textbf{21.42\%} \\
    & & \text{impv.} & +2.18\%  & +2.50\%  & +0.63\%  & +1.14\%  & +0.90\%  & +2.47\%  & +0.40\%  & +1.46\%  \\ \cmidrule{2-11}
    & \multirow{3}{*}{LoRA} & Vanilla & 29.78\% & 10.50\% & 14.14\% & \textbf{8.87\%} & 14.40\% & \textbf{21.31\%} & 9.50\% & 15.50\%\\
    & & CAT & \textbf{30.02\%} & \textbf{12.00\%} & \textbf{21.17\%} & 8.64\% & \textbf{14.58\%} & 20.82\% & \textbf{9.80\%} & \textbf{16.72\%} \\
    & & \text{impv.} & +0.24\% & +1.50\% & +7.03\% & -0.23\% & +0.18\% & -0.49\% & +0.30\% & +1.22\%\\ \hline

    \multirow{6}{*}{Qwen2.5-1.5B} & \multirow{3}{*}{Full} & Vanilla & 67.80\% & 51.00\% & 80.39\% & 45.34\% & 64.02\% & 79.52\% & 49.50\% & 62.51\% \\
    & & CAT & \textbf{69.73\%}& \textbf{56.00\%} & \textbf{83.33\%} & \textbf{47.08\%} & \textbf{64.79\%} & \textbf{82.18\%} & \textbf{52.10\%} & \textbf{65.03\%} \\
    & & \text{impv.} & +1.93\% & +5.00\% & +2.94\% & +1.74\% & +0.77\% & +2.66\% & +2.60\% & +2.52\% \\ \cmidrule{2-11}
    & \multirow{3}{*}{LoRA} & Vanilla & 74.33\% & 64.50\% & 67.89\% & 47.23\% & \textbf{70.52\%} & 89.88\% & 59.50\% & 67.69\% \\
    & & CAT & \textbf{76.27\%} & \textbf{65.00\%} & \textbf{69.87\%} & \textbf{50.04\%} & 68.58\% & \textbf{90.02\%} & \textbf{64.40\%} & \textbf{69.17\%} \\
   & & \text{impv.} & +1.94\% & +0.50\% & +1.98\% & +2.81\% & -1.94\% & +0.14\% & +4.90\% & +1.48\% \\ \hline

    \multirow{3}{*}{Llama-3.1-8B} & \multirow{3}{*}{LoRA} & Vanilla & 89.83\% & 72.00\% & 91.58\% & 65.66\% & 76.66\% & 90.94\% & 66.20\% & 78.98\%\\
    & & CAT & \textbf{90.31\%} & \textbf{72.50\%} & \textbf{91.84\%} & \textbf{66.57\%} & \textbf{78.51\%} & \textbf{91.33\%} & \textbf{69.70\%} & \textbf{80.11\%} \\
    & & \text{impv.} & +0.48\% & +0.50\% & +0.26\% & +0.91\% & +1.85\% & +0.39\% & +3.50\% & +1.13\% \\ \hline

  \end{tabular}
    }
  \caption{Performance comparison for different models and tasks. Full represents full parameters training. LoRA represents LoRA training."\dag" means training on GSM8K but testing on the given dataset.}
  \label{tab:model_comparison}

\end{table*}

\subsubsection{Expand to Downstream Tasks   }

We evaluated a broader range of downstream tasks. Under the in-domain setting, we tested four datasets related to mathematics and reasoning. Furthermore, we introduce an out-of-domain setting: models trained on GSM8K are evaluated on other math reasoning datasets. Although these datasets involve basic arithmetic reasoning, they differ in question formulation and answer formats, making them distributionally distinct and suitable for assessing cross-task generalization.   The results are shown in Table \ref{tab:model_comparison}. The CAT has shown consistent improvements across multiple settings and datasets. For example, under the full fine-tuning setting with Qwen, our method yields an average performance improvement of 2.52\%. Additionally, the CAT consistently outperforms the baseline in most OOD settings, demonstrating its stronger generalization ability. This suggests that guiding attention alignment toward human high-level causal reasoning can help models acquire deeper reasoning capabilities, rather than simply fitting to the training distribution.

\subsubsection{Ablation Studies}
\begin{figure}[h]
    \centering
    \includegraphics[width=1\linewidth]{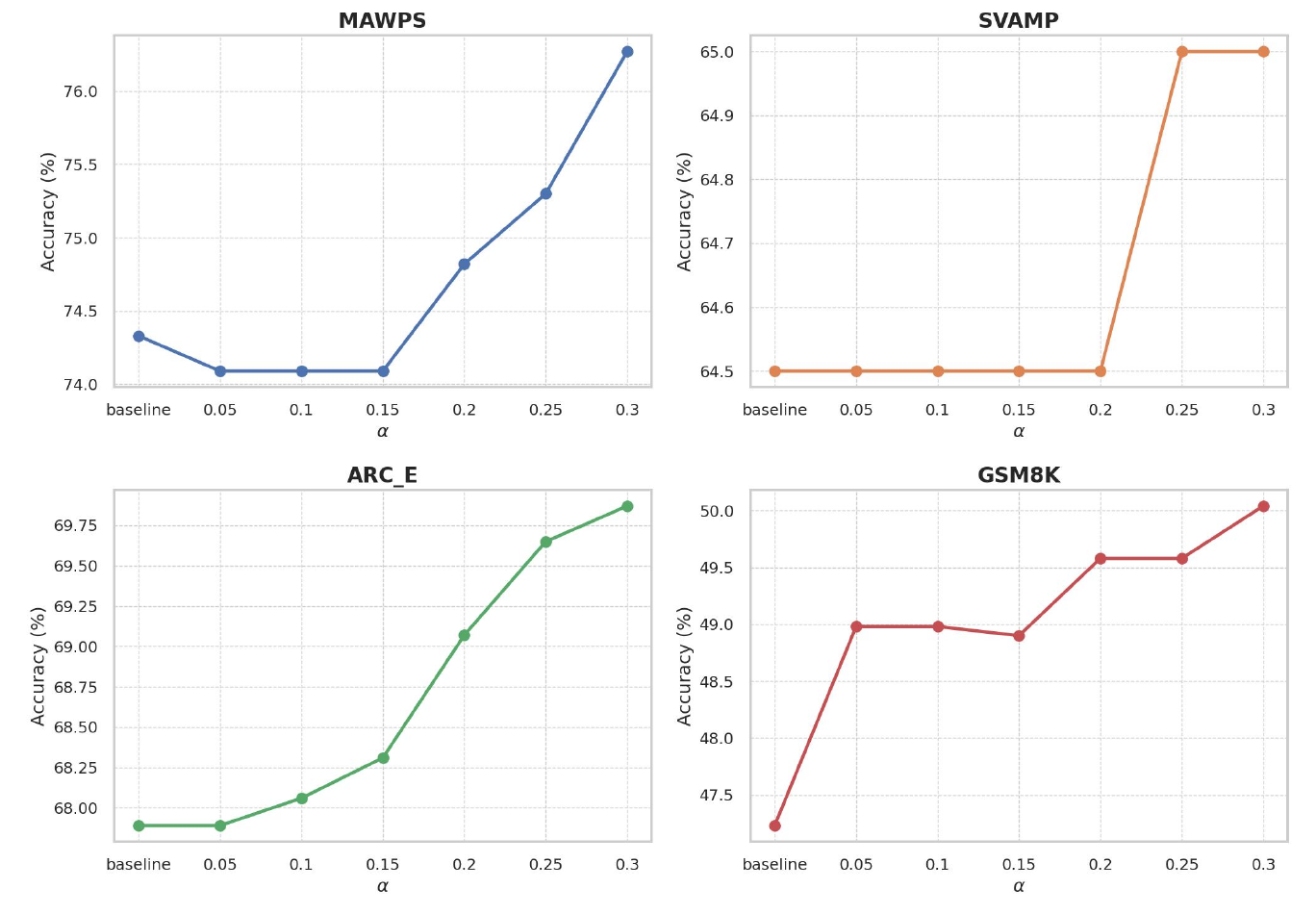}
    \caption{Performance under different $\alpha$ using LoRA.}
    \label{fig:xiaorongalpha}
\end{figure}

To explore the influence of the $\alpha$ and the $\gamma$, we conducted ablation experiments, taking Qwen2.5-1.5B as an example. We train LoRA and set $\alpha$ to 0.05, 0.1, 015, 0.2, 0.25, and 0.3. For $\gamma$, we set the coefficient $\gamma$ of $\mathcal{L}_{attn}$ as 1 (w/o $\gamma$). Using a weight decay strategy proves beneficial in most cases, as shown in Table \ref{tab:xiaoronggama}. As $\alpha$ increases, the performance gradually improves, shown in Figure \ref{fig:xiaorongalpha}. Through ablation experiments, we have demonstrated the effectiveness of each component of the CAT.

\begin{table}[h!]
\centering
\resizebox{0.5\textwidth}{!}{ 
\begin{tabular}{lccccc}
\toprule
\textbf{Mehod} & \textbf{MAWPS} & \textbf{SVAMP} & \textbf{ARC-E} & \textbf{GSM8K} & \textbf{Average} \\
\midrule
\text{CAT} & 69.73\%& \textbf{56.00\%} & \textbf{83.33\%} & \textbf{47.08\%} & \textbf{64.03\%} \\
\quad \text{w/o $\gamma$} & \textbf{71.91\%} & 54.50\% & 82.58\% & 45.64\% & 63.66\%  \\
\bottomrule
\end{tabular}
}
\caption{Ablation experiment on $\gamma$.}\label{tab:xiaoronggama}
\end{table}

\subsubsection{Cost Analysis and Powerful Assistant LLMs}

We replace ChatGLM-4-air with GPT-4o as the assistant LLMs, details in Appendix \ref{APP:Assistant LLMs}. Stronger assistant LLMs exhibit slightly better performance. Additionally, when using ChatGLM-4-air, the annotation cost can be as low as \$0.14\footnote{The pricing unit of the GLM API is in CNY, approximately 1 CNY per million tokens, which is equivalent to about 0.14 USD based on the exchange rate as of August 26, 2025.} per million tokens, in contrast to approximately \$18 with GPT-4o. Further details are provided in the Appendix \ref{APP:cost}.  Therefore, considering the cost, we recommend using GLM-4-air for better cost-effectiveness.

\section{Conclusions}

To address the issues of spurious correlations and the lack of causal knowledge in the inherent correlation-based Transformer, we propose CAT, a novel method that injects fine-grained causal knowledge during training. To evaluate the IID and OOD robustness, we introduce the STG benchmark. Extensive experiments across downstream tasks under diverse settings validate the positive impact of incorporating fine-grained causal knowledge and the \textit{Re-Attention} mechanism in downstream tasks. LLMs can effectively utilize causal knowledge for prediction and generation.

\section*{Limitations}

Due to resource limitations, we did not explore the performance impact of larger models, such as those exceeding 10B parameters (i.e., Qwen-2.5 14B), under the CAT method. Experimenting with larger LLMs could provide stronger insights. Additionally, CAT requires the introduction of an assistant LLM to label causal supervision signals, which will incur extra token overhead, although these costs remain within an acceptable range. This paper offers an empirical approach to causal knowledge injection. However, there is still significant room for exploration in terms of how to integrate causal knowledge into LLM mechanisms, starting from a more theoretical token-level causal modeling perspective. Due to the complexity of causal theory, even well-intentioned and fully professional humans may cause an LLM assistant to inject biases that do not exist in the training data. Causal relationships in the real world may be more complex, abstract, and context-dependent. The applicability of our approach to tasks that require a deeper and more nuanced understanding of causality has not yet been fully explored.

\section*{Ethical Concerns}

We declare that all authors of this paper acknowledge the \textit{ACM Code of Ethics} and honor the code of conduct. We do not foresee an immediate ethical or societal impact resulting from our work. However, our method provides opportunities for human experts to maliciously inject biases into LLMs, for example, by downplaying the causal effects of belonging to socially marginalized groups or by exaggerating the apparent correlation of spurious factors. Therefore, we urge users to exercise caution when using this method to avoid potential ethical and moral risks.

\section*{Acknowledgements}

This work was supported in part by the National Key Research and Development Program of China (2024YFE0203700), National Natural Science Foundation of China (62376243), "Pioneer" and "Leading Goose" R\&D Program of Zhejiang (2025C02037), and the Starry Night Science Fund of Zhejiang University Shanghai Institute for Advanced Study (SN-ZJU-SIAS-0010). All opinions in this paper are those of the authors and do not necessarily reflect the views of the funding agencies.

\bibliography{custom}

\appendix

\section{Data Generate Process}\label{DGP}

Machine learning theory posits that training and test sets are IID. However, due to the presence of spurious correlation, although the model's outcomes should be uniquely determined by causal features, the model may inadvertently capture these spurious correlations. This can lead to a reliance on spurious correlations rather than causal features when faced with a wide and diverse array of real-world scenarios, thereby compromising the model's reliability. So, for STG\_E, our data generation processing follows the formalized expression below, where in the IID scenario, it satisfies:

$$ \mathcal{C}^s_i, I^s_i \sim Rand(1,10)$$
$$ \mathcal{S}^s_i = r_i*\mathcal{C}_i^s$$
$$ f(\mathcal{C}^s)=\sum_i k_i*\mathcal{C}^s_i$$
\[
\mathcal{A}^{s} = 
\begin{cases} 
High, &  f(\mathcal{C}^s) \geq \mu_h \\
Low, & else\\
\end{cases}
\]
where $ r_i,k_i,\mu_h $ are hyperparameters to control the ratio of high risk and low risk. The accuracy of random guessing is $50\%$.

In the OOD scenario, the three elements are independent of each other:
$$ \mathcal{S}^{ood}_i, \mathcal{C}^{ood}_i, I^{ood}_i \sim Rand(1,10)$$

A specific example is as follows:

\begin{tcolorbox}[colframe=blue!50!black, colback=yellow!10, coltitle=black, fonttitle=\bfseries, sharp corners=southwest]
\textbf{Question:}
Here is the statistical data for a person. Please predict the probability of cancer. \\
Yellow fingers: 3, Weight: 1, Room size: 4, Certain gene: 4, Clothing size: 1, Smoking: 2, Hormones: 2, Exercise: 5
Here is the statistical data for a person. Please predict the probability of cancer. \\
\textbf{Answer:} Low Risk
\end{tcolorbox}

Specifically, the value of yellow fingers is 1.5 times that of smoking, the value of clothing size is the same as weight, and the value of hormones is 0.5 times that of exercise. All values are rounded down.
$\mu_h=7.2$ and 
\begin{multline}
f(\mathcal{C}^s) = 1.2*\#Smoking+0.7*\#
 Weight\\ 
 -\#Exercise \nonumber
 \end{multline}

\begin{figure}[h]
    \centering
    \includegraphics[width=1\linewidth]{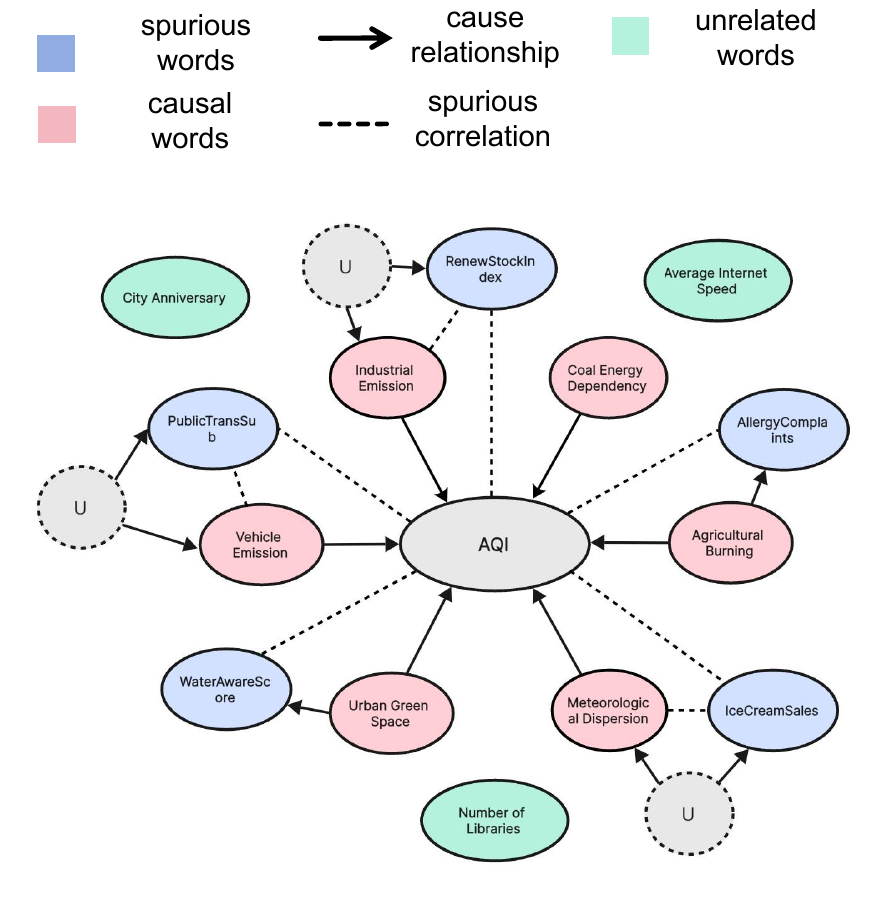}
    \caption{The causal graph underlying the data generation process of the STG\_H dataset.}
    \label{fig:cg}
\end{figure}

For the STG\_H dataset, we follow the causal graph shown in Figure \ref{fig:cg}. The process is similar to that of STG\_E, and the implementation details can be found in our code repository. The specific differences among the subsets of the STG dataset are shown in Table \ref{tab:stg-subsets}.

\begin{table*}[ht]
\centering
\begin{tabular}{c c c c c c c}
\toprule
\textbf{Name} & \textbf{Subset} & \textbf{Training Size} & \textbf{Test Size} & \textbf{OOD Test Size} & \textbf{Node Number} & \textbf{Answer} \\
\midrule
STG\_E & STG\_S & 0.4k & 0.4k & 0.4k & 8 & high/low risk \\
STG\_E & STG\_M & 0.8k & 0.4k & 0.4k & 8 & high/low risk \\
STG\_E & STG\_L & 1.6k & 0.4k & 0.4k & 8 & high/low risk \\
STG\_H & --     & 3k   & 1k   & 1k   & 14 & 0-100 \\
\bottomrule
\end{tabular}
\caption{Details of STG dataset subsets.}
\label{tab:stg-subsets}
\end{table*}

\section{Prompt Template}\label{Prompt Template}
In different downstream tasks, we manually crafted prompt templates to guide the assistant LLM in extracting token-level causal associations. The prompt templates for different tasks are as follows:

\newtcolorbox[auto counter, number within=section]{mybox}[2][]{colframe=blue!30!gray, colback=yellow!10, 
coltitle=black, fonttitle=\bfseries, title={\tikz[baseline]{\node[yellow!20, text=black, anchor=base] {\textbf{#2}};}},
sharp corners=southwest, breakable, #1}

\begin{mybox}[title={SVAMP}]

You need to evaluate the causal importance relationships between tokens in text data from the field of mathematical reasoning. Among them, entities, values, and keywords containing operation symbols are crucial for numerical reasoning. The data is used to train autoregressive models, so tokens that appear later can only see the tokens that come before them. Please output the important tokens for executing mathematical reasoning tasks during training, along with the tokens they should focus on from the preceding context as causal associations (which can be more than one). Present the output JSON string in a dict format, such as \{"A":[...], "B":[...],...\}. You should only output JSON without other contents. Note that the Answer part is considered important and must be analyzed.

\textbf{\#\#demo}

If they are already at 659 feet and the cave is 762 feet deep. How much farther until they reach the end of the cave?\textbf{Answer:} 103.0

\textbf{\#\#output}

\{\\
"762 feet deep":["the cave"],\\
"until":["How much farther"],\\
"Answer":["659 feet","762 feet", "until", "end of the cave"],\\
"103.0":["659 feet", "and", "762 feet", "Answer"]\\
\}

\#\#Please output following sentence importance between tokens. The final answer at the end and the corresponding number's importance must always be analyzed (such as 103.0 shown above).
\end{mybox}

\begin{mybox}[title={ARC\_E}]

You need to evaluate the causal importance relationships between tokens in text data from the field of reasoning. You only need to consider the tokens that have the greatest impact on the final answer. The data is used to train autoregressive models, so tokens that appear later can only see the tokens that come before them. Please output the important tokens for executing reasoning tasks during training, along with the tokens they should focus on from the preceding context as causal associations (which can be more than one). Present the output JSON string in a dict format, such as \{"A":[...],"B":[...],...\}. Note that the Answer part is considered important and must be analyzed.
Below I will give you a single-choice question. You need to analyze the most important part of each option for the answer, and together with the answer, form the causal relationship that needs to be considered to generate the answer. Note that only the token behind can notice the previous word, and keep the autoregressive characteristics, such as "option content": "option A/B/C/D". The specific example is as follows:

\textbf{\#\#demo:}

Which factor will most likely cause a person to develop a fever?

A. a leg muscle relaxing after exercise

B. a bacterial population in the bloodstream

C. several viral particles on the skin

D. carbohydrates being digested in the stomach

Answer: B

\textbf{\#\#output:}

\{

"develop a fever":["factor","cause"],

"leg muscle relaxing":["A."],

"bacterial population":["B."],

"viral particles":["C."],

"digested in the stomach":["D."],

"Answer: B":["A.", "leg muscle relaxing", "B.","bacterial population", "C.","viral particles", "D.",  "digested in the stomach"]

\}

\#\#Please output following sentence importance between tokens. The final answer at the end and the corresponding number's importance must always be analyzed (such as Answer: B shown above). You should only output JSON string without other contents.

\end{mybox}

\begin{mybox}[title={GSM8k}]

You need to evaluate the causal importance relationships between tokens in text data from the field of mathematical reasoning. Among them, entities, values, and keywords containing operation symbols are crucial for numerical reasoning. The data is used to train autoregressive models, so tokens that appear later can only see the tokens that come before them. Please output the important tokens for executing mathematical reasoning tasks during training, along with the tokens they should focus on from the preceding context as causal associations (which can be more than one). Present the output JSON string in a dict format, such as \{"A":[...],"B":[...],...\}. You should only output JSON without other contents. Note that the Answer part is considered important and must be analyzed.

\textbf{\#\#demo}

Natalia sold clips to 48 of her friends in April, and then she sold half as many clips in May. How many clips did Natalia sell altogether in April and May? Answer: Natalia sold 48\/2 = <<48\/2=24>>24 clips in May. Natalia sold 48+24 = <<48+24=72>>72 clips altogether in April and May.\#\#\#\# 72

\textbf{\#\#output}

\{

"in April":["48"],

"in May": ["half as many clips", "48\/2 = <<48\/2=24>>24 clips", "48"],

"72 clips": ["How many clips", "sell altogether", "48+24", "in April", "in May"],

"\#\#\#\# 72":["How many clips","in April and May","48+24","72 clips"]

\}

\#\#Please output following sentence importance between tokens. The final answer at the end and the corresponding number's importance must always be analyzed (such as \#\#\#\# 72 shown above). Please try to use the most refined causal characteristics to summarize the causal process of the answer

\end{mybox}

\begin{mybox}[title={MAWPS}]

You need to evaluate the causal importance relationships between tokens in text data from the field of mathematical reasoning. Among them, entities, values, and keywords containing operation symbols are crucial for numerical reasoning. The data is used to train autoregressive models, so tokens that appear later can only see the tokens that come before them. Please output the important tokens for executing mathematical reasoning tasks during training, along with the tokens they should focus on from the preceding context as causal associations (which can be more than one). Present the output JSON string in a dict format, such as {"A":[...],"B":[...],...}. You should only output JSON without other contents. Note that the Answer part is considered important and must be analyzed.

\textbf{\#\#demo}

William has 2 bottle caps. He buys 41 more. How many bottle caps does William have in all? Answer: 43.0

\textbf{\#\#output}

\{

"2 bottle caps": ["William"], 

"41 more": ["He buys"],

"William have": ["How many bottle caps"],

"Answer": ["How many bottle caps"],

"43.0": ["2 bottle caps", "41 more"]

\}

\#\#Please output following sentence importance between tokens. The final answer at the end and the corresponding number's importance must always be analyzed (such as 43.0 shown above).
\end{mybox}

\section{Details of the Experimental Results on the STG Dataset}\label{APP:STG}

\begin{figure*}[t]
    \centering
    \includegraphics[width=1\linewidth]{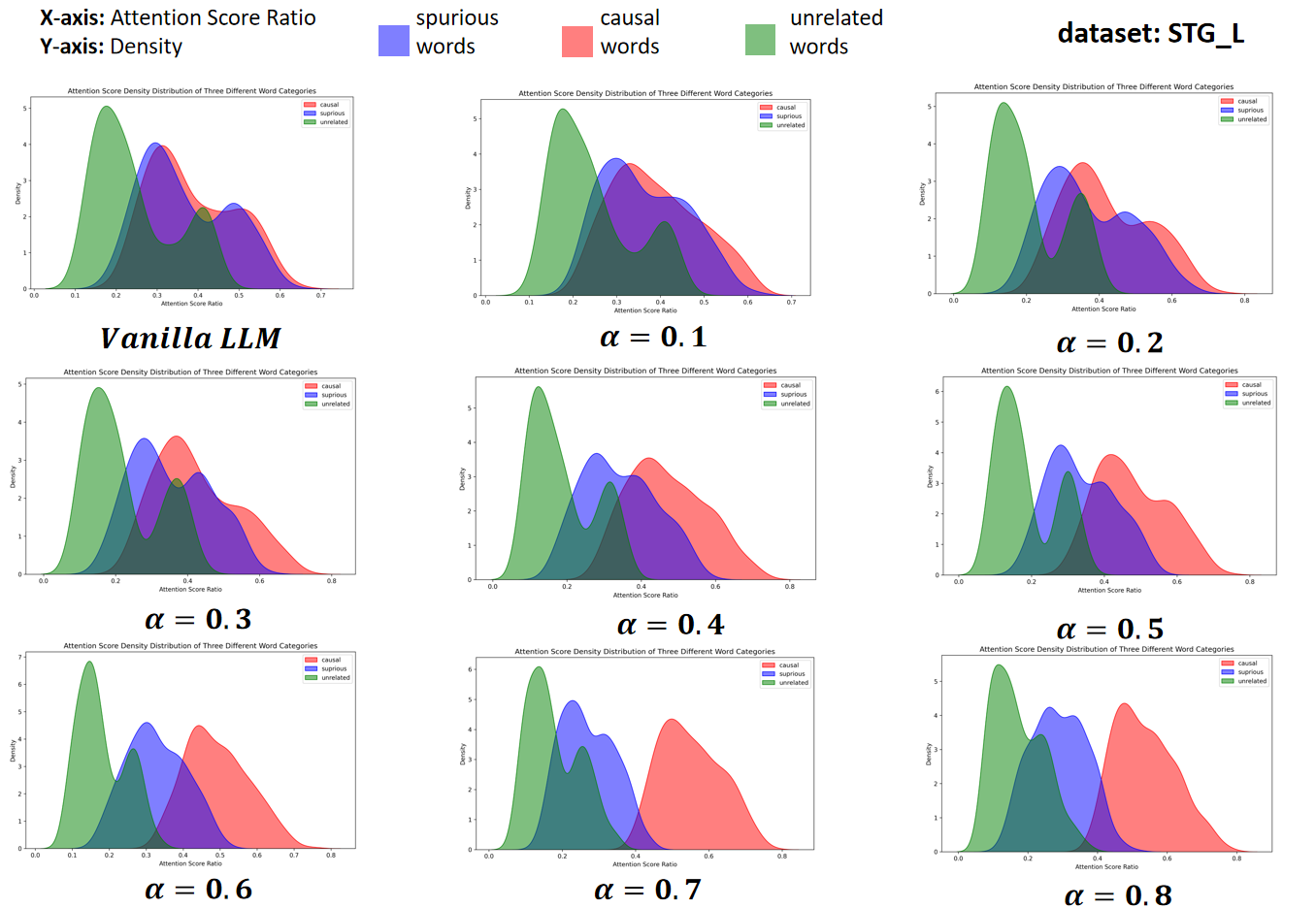}
    \caption{Visualization of the attention distribution on the STG\_L dataset.}
    \label{fig:STG_L}
\end{figure*}

\begin{figure*}
    \centering
    \includegraphics[width=1\linewidth]{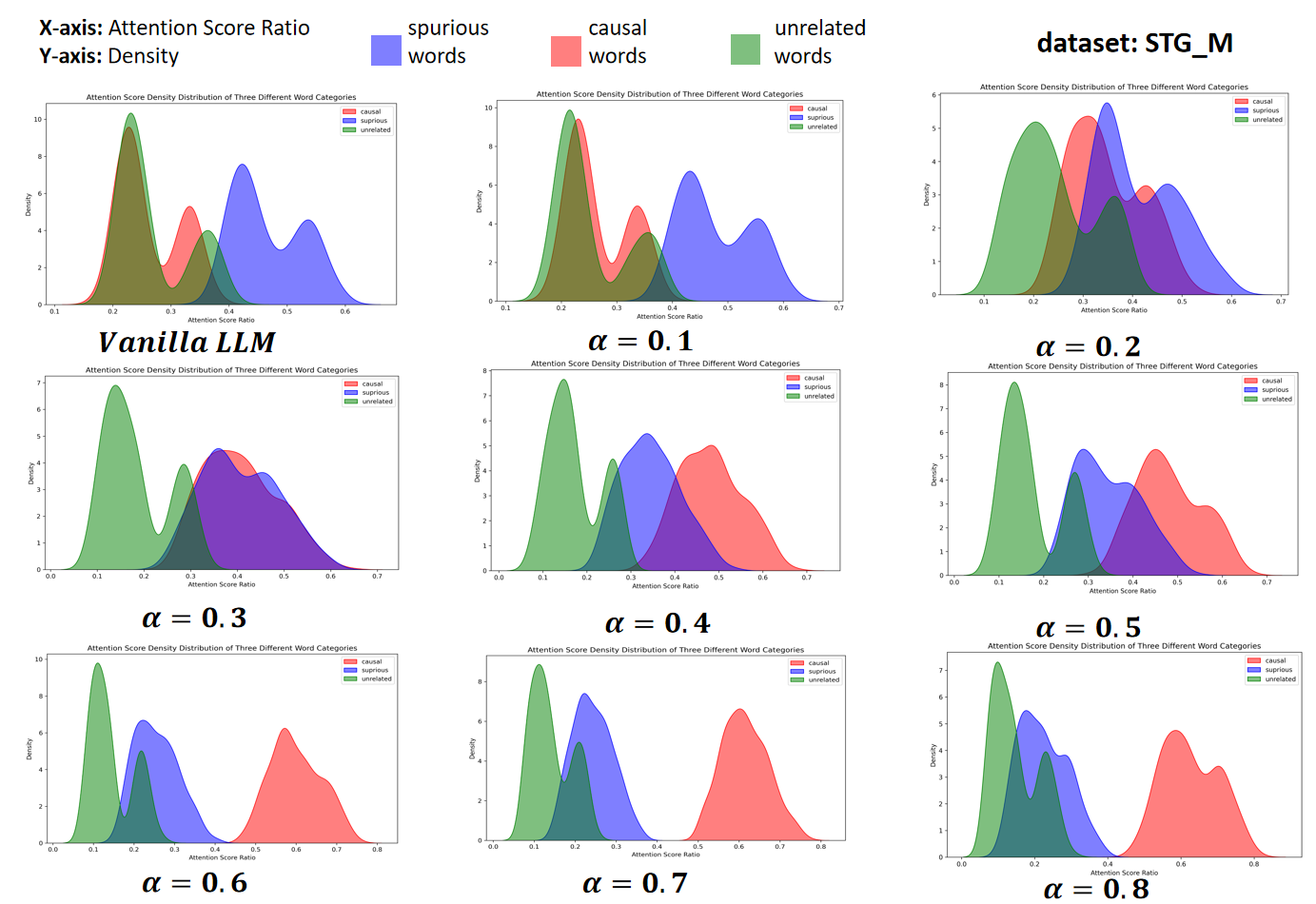}
    \caption{Visualization of the attention distribution on the STG\_M dataset.}
    \label{fig:STG_M}
\end{figure*}

\begin{figure*}
    \centering
    \includegraphics[width=1\linewidth]{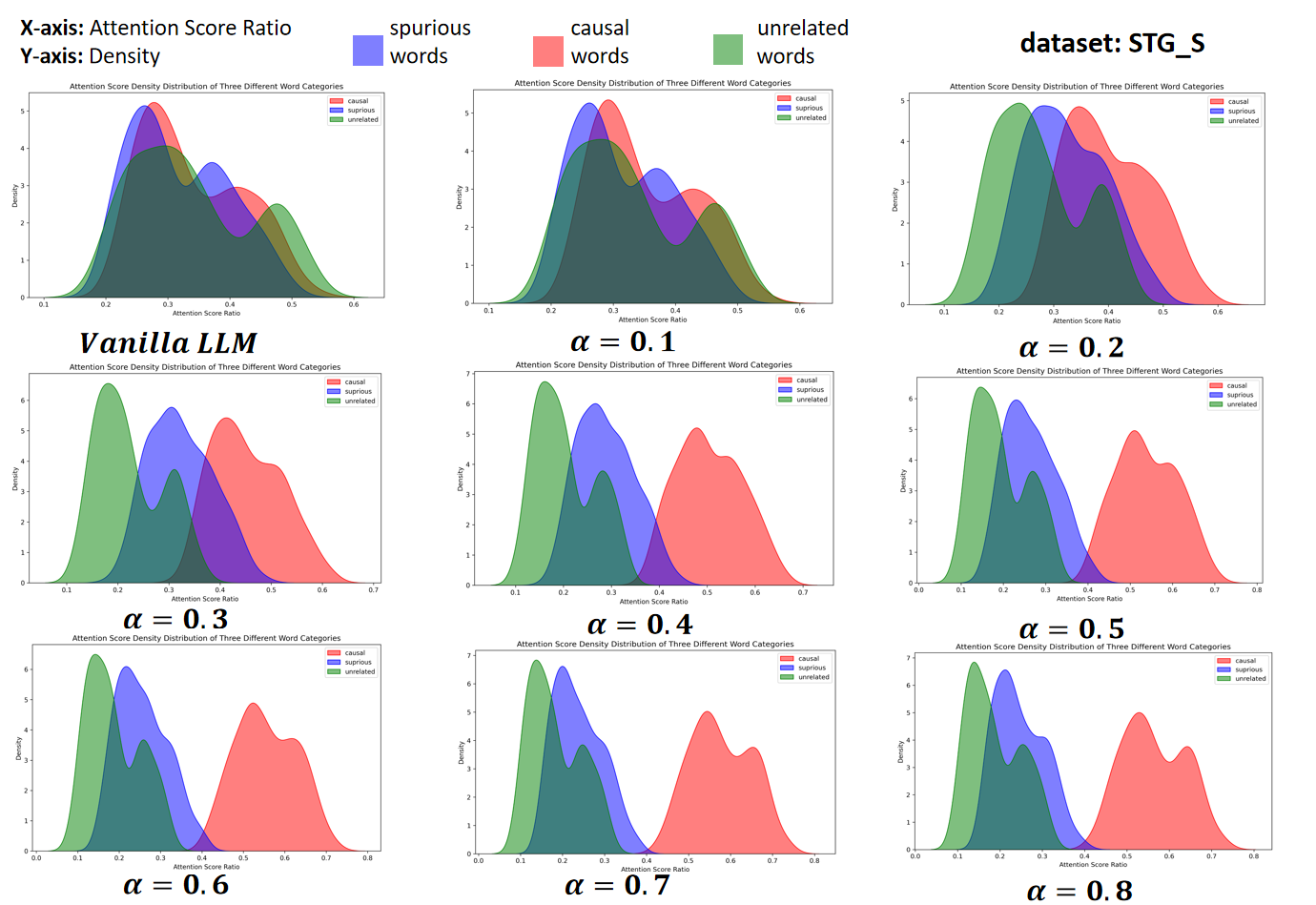}
    \caption{Visualization of the attention distribution on the STG\_S dataset.}
    \label{fig:STG_S}
\end{figure*}

We visualized the distribution of attention scores for STG under different $\alpha$ under TinyLlama-1.1B, as shown in Figure \ref{fig:STG_L}, \ref{fig:STG_S}, \ref{fig:STG_M}.

As we can see, as $\alpha$ increases, the model's attention to causal words gradually strengthens, and within a certain range, both IID and OOD performance improve consistently.

\begin{figure}[h]
    \centering
\includegraphics[width=1\linewidth]{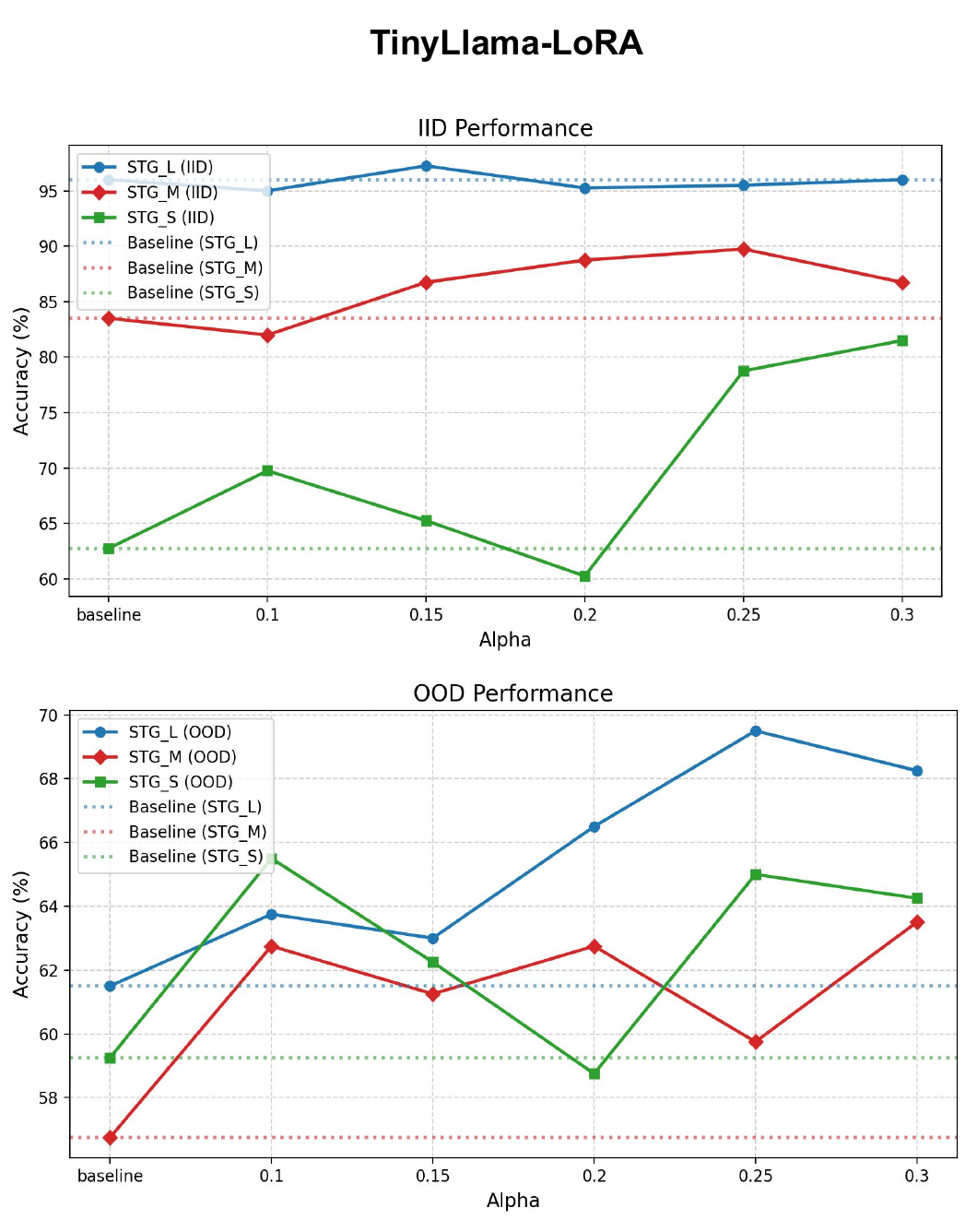}
    \caption{Performance of STG dataset under different $\alpha$ using TinyLlama-1.1B-LoRA.}
    \label{fig:TinyLlama-Lora}
\end{figure}

In addition, we analyzed the performance changes of LoRA setting under different $\alpha$, as shown in Figure \ref{fig:TinyLlama-Lora}. 

\section{Cost Analysis}\label{APP:cost}

To estimate annotation costs, we randomly sampled 10 instances from the GSM8K dataset and annotated them using GPT-4o. The statistics for these samples are as follows: the average token length of the original inputs was 168.4, the average length of the prompts (including instructions for causal supervision signal extraction) was 570.0, and the average length of the model's output completions was 163.9.

Based on the official pricing of GPT-4o, we estimate the maximum additional cost per 1 million tokens annotated, assuming no cache hits, as:

$$
\left(\frac{570.0}{168.4} \times \$2.50\right) + \left(\frac{163.9}{168.4} \times \$10.00\right) \approx \$18.19
$$

For comparison, we also consider the use of ChatGLM-4-air, which was employed in our experiments. Given that the output lengths across different LLMs are relatively consistent, we adopt the same average prompt and output lengths as a reasonable approximation. Based on the official pricing of ChatGLM-4-air, the estimated cost is:

$$
\frac{570.0 + 163.9}{168.4} \times 0.25 \approx 1.09 (CNY)
$$

These results demonstrate the low cost of our approach. By utilizing batch API calls to proprietary large language models, our method enables efficient large-scale data annotation. Moreover, it is compatible with mainstream closed-source models (e.g., GPT-4, ChatGLM, Gemini), making it adaptable to various application requirements.

\section{Comparison of Different Assistant LLMs}\label{APP:Assistant LLMs}

To fairly compare the performance of different assistant LLMs. We use GPT4o to replace GLM-4-air and conduct comparative experiments in MAWPS. Specifically, we conducted six experiments on the $\alpha$ grid of 0.05, 0.1, 0.15, 0.2, 0.25, and 0.3, and reported the average and max accuracy. The experimental results are shown in Table \ref{tab:model_perf}. Stronger teacher models exhibit slightly better performance. Considering the cost, we recommend using ChatGLM-4-air.

\begin{table}[h!]
\centering
\resizebox{0.5\textwidth}{!}{ 
\begin{tabular}{lcccc}
\toprule
\textbf{Model} & \textbf{Setting} & \textbf{Source} & \textbf{Avg ± Std} & \textbf{Max} \\
\midrule
\multirow{4}{*}{Tinyllama-1.1B}& \multirow{2}{*}{Full} & GPT
  & 0.4165 ± 0.0143 & 0.4310  \\
 & & GLM  & \textbf{0.4185 ± 0.0216} & \textbf{0.4552}  \\ \cmidrule{2-5}
&\multirow{2}{*}{Lora} 
   & GPT & \textbf{0.2986 ± 0.0023} & \textbf{0.3027} \\
   & & GLM  & 0.2982 ± 0.0026 & 0.3002 \\
\midrule

\multirow{4}{*}{Qwen2.5-1.5B}& \multirow{2}{*}{Full} & GPT
  & \textbf{0.7288 ± 0.0157} & \textbf{0.7482}  \\
 & & GLM-4  & 0.7123 ± 0.0099 & 0.7264  \\ \cmidrule{2-5}
&\multirow{2}{*}{Lora} 
   & GPT & \textbf{0.7502 ± 0.0055} & 0.7579 \\
   & & GLM  & 0.7478 ± 0.0081 & \textbf{0.7627} \\
\midrule

\multirow{2}{*}{Llama-3.1-8B} & \multirow{2}{*}{Lora} 
   & GPT & 0.8979 ± 0.0055 & \textbf{0.9055}\\
   & & GLM-4  & \textbf{0.9003 ± 0.0047} & \textbf{0.9055} \\
\bottomrule
\end{tabular}
}
\caption{The impact of different assistant LLMs on performance. GPT represents GPT-4o, and GLM represents ChatGLM-4-air.}
\label{tab:model_perf}
\end{table}

\section{Details of the Hyperparameters Used During Training.}\label{APP:hyper}

All comparisons between baselines and the CAT are obtained after sufficient training with the same hyperparameter settings.  Due to differences in tasks, models, and GPU memory limitations, we ensure that the product of batch size and gradient accumulation steps remains consistent across different downstream tasks under the same model setting. For testing, we use greedy decoding with the following parameter settings, as shown in Table \ref{tab:test_hyperparams}.

\begin{table}[h]
\centering
\begin{tabular}{ll}
\hline
\textbf{Parameter} & \textbf{Value} \\
\hline
max\_new\_tokens & 512 \\
batchsize & 64 \\
num\_return\_sequences & 1 \\
do\_sample & False \\
\hline
\end{tabular}
\caption{Hyperparameters used during testing}
\label{tab:test_hyperparams}
\end{table}

All random seeds used in this paper are set to 42 to ensure the reproducibility of the experiments. For more experimental details, please refer to our code repository. 

Due to differences in attention distributions across different models and tasks, we observe empirically that setting the $\alpha$ parameter between 0.05 and 0.35 yields better results. We strive to keep the $\alpha$ hyperparameter consistent within the same model whenever possible. In the five reasoning downstream tasks, we report the test results based on the best-performing model on the validation set. For TinyLlama-1.1B LoRA, we set $\alpha$ to 0.2. For Qwen-2.5-1.5B, both LoRA and the full model, we use 0.3. For TinyLlama-1.1B, we use two settings: 0.15 and 0.2. For Llama-3.1-8B-Instruct, we use 0.25 and 0.3. For STG\_H, we uniformly set the $\alpha$ parameter to 0.3 for all LoRA models and 0.2 for all full-parameter models. For STG\_E, we perform a grid search with an interval of 0.05 in the range from 0.05 to 0.35. Specifically, for the TinyLlama-1.1B full-parameter model on STG\_E, we set $\alpha$ to 0.6. The $\alpha$ parameter determines the degree to which the model relies on causal associations. How to efficiently identify the optimal $\alpha$ value is left for future work.

For the STG\_E dataset, we apply LoRA fine-tuning to TinyLlama and Qwen using a learning rate of $1\times e^{-3}$, and full fine-tuning on Qwen with a learning rate of $1\times e^{-4}$.

\section{Use Of AI Assistants}
We used generative AI, ChatGPT, to check for syntactic and grammatical errors in the manuscript. We carefully verified the correctness of the revised content.

\end{document}